\pdfoutput=1
\documentclass[10pt, a4paper]{article}
\usepackage{lrec}
\usepackage{multibib}
\newcites{languageresource}{Language Resources}
\usepackage{graphicx}
\usepackage{tabularx}
\usepackage{soul}
\usepackage{titlesec}
\titleformat{\section}{\normalfont\large\bf\center}{\thesection.}{1em}{}
\titleformat{\subsection}{\normalfont\SmallTitleFont\bf\raggedright}{\thesubsection.}{1em}{}
\titleformat{\subsubsection}{\normalfont\normalsize\bf\raggedright}{\thesubsubsection.}{1em}{}
\renewcommand\thesection{\arabic{section}}
\renewcommand\thesubsection{\thesection.\arabic{subsection}}
\renewcommand\thesubsubsection{\thesubsection.\arabic{subsubsection}}

\usepackage{epstopdf}
\usepackage[utf8]{inputenc}

\usepackage{hyperref}
\usepackage{xstring}

\usepackage{color}

\usepackage{booktabs}
\usepackage{multirow}

\title{Subtitles to Segmentation: Improving Low-Resource Speech-to-Text Translation Pipelines}

\name{David Wan$^{1}$, 
    Zhengping Jiang$^{1}$, 
    Chris Kedzie$^{1}$, 
    Elsbeth Turcan$^{1}$,\\
    {\bf \large Peter Bell$^{2}$
    and
    Kathleen McKeown$^{1}$}
    }

\address{$^{1}$Columbia University, $^{2}$University of Edinburgh \\
 \{dw2735, zj2265\}@columbia.edu, 
 \{kedzie,eturcan,kathy\}@cs.columbia.edu,
 peter.bell@ed.ac.uk\\}

\abstract{
In this work, we focus on improving ASR output segmentation in the context
of low-resource language speech-to-text translation. ASR output segmentation is crucial, as
ASR systems segment the input audio using purely acoustic information and are not guaranteed to output sentence-like segments. Since most MT systems expect sentences as input, feeding in longer unsegmented passages can lead to sub-optimal performance.
We explore the feasibility of using datasets of subtitles from TV shows and movies to train better ASR
segmentation models. 
We  further  incorporate  part-of-speech  (POS)  tag  and  dependency label information (derived from the unsegmented ASR outputs) into our segmentation model. We show that this noisy syntactic information can improve model accuracy.
We evaluate our models intrinsically on segmentation quality  and  extrinsically  on  downstream  MT  performance, as well as downstream tasks including cross-lingual information retrieval (CLIR) tasks and human relevance assessments. Our model shows improved performance on downstream tasks for Lithuanian and Bulgarian.
\newline \Keywords{Speech Segmentation, Lithuanian, Bulgarian, Low-Resource Languages } }

\begin{document}

\maketitleabstract

\section{Introduction}

A typical pipeline for speech-to-text translation (STTT) uses a cascade of automatic speech recognition (ASR), ASR output segmentation, and machine translation (MT) components \cite{segmentationimportance}. ASR output segmentation is crucial, as
ASR systems segment the input audio using purely acoustic information and are not guaranteed to output sentence-like segments
(i.e., one utterance may be split if the speaker pauses in the middle, or utterances may be combined if the speaker does not pause).
Since most MT systems expect sentences as input, feeding in longer unsegmented passages can lead to sub-optimal performance \cite{koehn2017}.

When the source language is a low-resource language, suitable training data may be very limited for ASR and MT, and even nonexistent for
segmentation. Since typical low-resource language ASR audio datasets crawled from the web do not have hand-annotated segments we propose deriving proxy segmentation datasets from TV show and movie subtitles. Subtitles typically contain boundary information like sentence-final punctuation and speaker turn information, even if they are not exact transcriptions.
\begin{figure*}
    \centering

    \noindent\fbox{%
    \parbox{\textwidth}{%
        I think you should know something. You know. I ... 
    \small  \center
    $\begin{array}{ccccccccccccccccccc}
        y& & = & 0 & 0 & 0 & 0 & 0 & 1 & 0 & 1 & 0  &\cdots \\
    & token & = & i & think & you & should & know & something & you & know & i  & \cdots \\
    x&dep & = & nsubj & root & nsubj & aux & ccomp & obj & nsubj & acl:relcl & nsubj  &\cdots \\
    &pos & = & PRON & VERB & PRON & AUX & VERB & PRON & PRON & VERB & PRON   &\cdots 
    
    \end{array}
    $
    }%
}
    \caption{An excerpt of subtitles (top) and the corresponding segmentation data derived from it (bottom). Punctuation is to mark boundaries $y_i=1.$ Part-of-speech and dependency relations are parsed from each document.}
    \label{fig:tv-example}
\end{figure*}

We further incorporate part-of-speech (POS) tag and dependency label information (derived from the unsegmented ASR outputs) into our segmentation model. 
This noisy syntactic information %
can improve model accuracy.

We evaluate our models intrinsically on segmentation quality and extrinsically on downstream MT performance.
Since the quality of the underlying MT of low-resource languages is relatively weak,
we also extrinsically evaluate our improved STTT pipeline on document and 
passage-level
cross-lingual information retrieval (CLIR) tasks. We report results for
two translation settings: Bulgarian (BG) to English and Lithuanian (LT) to English.

This paper makes the following contributions: (i) We propose the use of subtitles as a proxy dataset for ASR segmentation. 
(ii) We develop a simple neural tagging model using noisy syntactic features on this dataset.
(iii) We show downstream performance increases on several extrinsic tasks: MT and document and passage-level CLIR tasks.

\section{Related Work}
Segmentation in STTT has been studied quite extensively in high resource settings. Earlier models use kernel-based SVM models to predict sentence boundaries from ngram and part-of-speech features derived from a fixed window size \cite{Sridhar2013SegmentationSF}.

Recent segmentation models use neural architecture, such as LSTM \cite{kit2018} and Transformer models \cite{kit2019}. These models benefit from large training data available for the high-resource languages. For example, the STTT task for English audio to German include TED corpus, which contains about 340 hours of well transcribed data. To our knowledge, these data do not exist for the languages we are interested in. In addition, these models predict full punctuation marks as well as casing for words (binary classification of casing). However, since our translation models are trained on unpunctuated texts, we restrict the classification task to predicting full stop boundaries only.

Although recent works have looked at end-to-end speech-to-text translation, in a high-resource setting, these models \cite{Vila2018EndtoEndST} achieved at most a 0.5 BLEU score improvement over a weak cascaded model. In general, the available data for end-to-end neural models is insufficient or non-existent in all but the most specific circumstances; for any pair of languages there will inevitably be far less translated speech data available than (a) monolingual transcribed speech data; (b) monolingual language modelling training data; or (c) parallel corpora of translated text data.  This means that separate ASR and MT systems will generally have the benefit of training on much larger datasets.
  
\begin{table}
\centering
\begin{tabular}{c cc|cc}
\toprule
\multirow{2}{*}{Corpus}       & \multicolumn{2}{c}{BG} & \multicolumn{2}{c}{LT} \\
& P & U &P & U \\
\midrule
OpenSub. & 164,798   & 41.9   & 32,603  & 49.5\\
ANALYSIS & 215 & 37.3 & 312 & 57.2 \\
DEV   &  238 & -- & 258 & --\\
\bottomrule
\end{tabular}
\caption{Number of passages (P) in each dataset and average number of utterances 
per passage (U).}
\label{segdat}
\end{table}
\begin{table}
\centering
\begin{tabular}{ cccc}
\toprule
Lang.  & Model & F1 $\uparrow$ %
    & WD $\downarrow$ \\
\midrule
 \multirow{2}{*}{BG}  &  Sub &  \textbf{56.78} %
 & \textbf{33.9*}\\
  & Sub+S  & 56.40 %
  & 34.4\\
  \midrule
 \multirow{2}{*}{LT}  & Sub   & 44.14 %
            & 49.2\\
                        & Sub+S & \textbf{45.94*} %
                        & \textbf{47.0*}\\
\bottomrule
\end{tabular}
\caption {Intrinsic evaluation of F1 and windowdiff(WD) on ANALYSIS data. +S indicates 
models with syntactic features. * indicates statistical significance}
\label{intrinsic}
\end{table}

\section{Datasets}
  
  \subsection{Segmentation Datasets} \label{sec:seg-data}

  We obtain BG and LT subtitles from the OpenSubtitles 2018 corpus \cite{lison2016opensubtitles2016}, which contains monolingual subtitles for 62 languages drawn from movies and television. We sample 10,000 documents for BG and all available documents for LT (1,976 in total).
  Sentences within a document are concatenated together. Some documents are impractically long and do not match our shorter evaluation data, so we divide each document into 20 equally sized passages (splitting on segment boundaries), roughly matching the average evaluation document size.
  In addition to speaker turns in subtitles, we treat any of the characters \emph{():-!?.} as segment boundaries. 
   We  split the data into a training (75\%) and validation set.
   See \autoref{segdat} for corpus statistics.
  
  \subsection{Speech Datasets}

  To perform extrinsic evaluation of a STTT pipeline, we use the speech collections from the MATERIAL\footnote{\url{www.iarpa.gov/index.php/research-programs/material}} program,
  which aims at finding relevant audio and text documents in low resource languages given English queries. This can be framed as an cross-language information retrieval (CLIR) task, where STTT plays a crucial part in improving the quality of downstream tasks of machine translation and information retrieval.
  
  The speech data consists of three domains (news broadcast (NB), topical broadcast (TB) such as  podcasts, and conversational speech (CS)) from multiple low-resource languages. NB and TB have one speaker and are more formal, while CS has two and is more casual. For each language, we have two collections of speech documents, the ANALYSIS and DEV sets (each containing a mix of NB, TB, and CS).
  Only the ANALYSIS datasets include ground truth transcriptions (including segmentation), allowing us to evaluate segmentation and translation quality. However, we can use both datasets for the extrinsic CLIR evaluation since MATERIAL provides English queries with ground truth relevance judgements.

\section{Segmentation Model}

We treat ASR  segmentation as a sequence tagging problem. Let $x_1, \ldots, x_n \in \mathcal{V}^n$ be a passage of $n$ ASR output tokens drawn from a finite vocabulary $\mathcal{V}$.
We also define an indicator variable $y_i$ for each token,
where $y_i=1$ indicates %
a segment boundary 
between tokens $x_{i}$ and $x_{i+1}$. Each token $x_i$ is additionally associated with a corresponding 
POS tag and dependency label. An example input and output are shown in \autoref{fig:tv-example}.

We explore a Long Short-Term Memory (LSTM)-based model architecture for this task.
In the input layer we represent each word as a 256-dimensional word embedding; when using syntactic information, we also concatenate its POS tag and dependency label embeddings (both 32-dimensional).
POS tags and dependency labels are obtained using the UDPipe 2.4 parser \cite{udpipe2017}.
Since we do not have punctuation on actual ASR output,
we parse each document with this information removed.
Conversational speech between two speakers comes in separate channels for each speaker so we concatenate the output of each channel and treat it as a distinct document when performing segmentation. The segmentation are then merged back into one document 
using segmentation timestamp information before being used in downstream evaluations.

We then apply a bi-directional LSTM to the input sequence of 
embeddings to obtain a 
sequence of $n$ hidden states, each of 256 dimensions 
(after concatenating the output of each direction). 
Each output state is then passed through a linear projection layer with logistic sigmoid output to compute the  probability of a segment boundary 
$p(y_i=1|x)$. The log-likelihood of a single passage/boundary annotation
pair is $\log p(y|x) = \sum_{i=1}^n \log p(y_i|x)$. All embeddings and parameters are
learned by minimizing the
negative log-likelihood on the training data using stochastic gradient
descent.

\begin{table*}[ht]
\centering
\begin{tabular}{ llccc|ccc|ccc }
\toprule
\multirow{2}{*}{Lang.} & \multirow{2}{*}{Model} &\multicolumn{3}{c}{EDI-NMT} & \multicolumn{3}{c}{UMD-NMT}  & \multicolumn{3}{c}{UMD-SMT} \\
 & & NB & TB & CS & NB & TB & CS & NB & TB & CS\\
\midrule
\multirow{3}{*}{BG} & Acous.   & 24.49 & 24.65 & 7.13 & \textbf{33.25} & 29.82 & 10.32 & \textbf{35.30} & 31.11 & 11.08\\
& Sub & 24.83 & \textbf{25.28} & \textbf{8.07} & 32.89 & \textbf{30.35} & 11.10 & 35.15 & \textbf{31.55}  & 11.32\\
& Sub+S & \textbf{24.90} & 25.25 & 8.04 & 32.96 & 30.23 & \textbf{11.24} & 35.16 & \textbf{31.55} & \textbf{11.57}\\
\midrule
\multirow{3}{*}{LT} & Acous. &   \textbf{16.03} & \textbf{17.00} & \textbf{6.53} & \textbf{16.31} & \textbf{18.67} & \textbf{5.92} & \textbf{16.52}  & \textbf{17.60} & \textbf{6.34} \\
& Sub &  14.83 & 15.59 & 6.33 & 15.41 & 17.47 & 4.66 & 15.93 & 17.14 & 5.86 \\
& Sub+S& 14.97 & 15.77 & 6.43 & 15.40 & 17.54 & 5.11 & 15.76 & 17.19 & 6.00 \\

\bottomrule
\end{tabular}
\caption {Document level BLEU scores on ANALYSIS set. +S indicates model with syntactic features.}
\label{bleu}
\end{table*}

\begin{table*}[ht]
\label{mqwvltopus}
\centering
\begin{tabular}{ llccc|ccc|ccc }
\toprule
 \multirow{2}{*}{Lang.} & \multirow{2}{*}{Model} & \multicolumn{3}{c}{EDI-NMT} & \multicolumn{3}{c}{UMD-NMT}  & \multicolumn{3}{c}{UMD-SMT} \\
 & & NB & TB & CS & NB & TB & CS & NB & TB & CS\\
\midrule
\multirow{3}{*}{BG} & Acous. & 0.289 & \textbf{0.482} & 0.052 & 0.394 & 0.175 & 0.005 & 0.426 & 0.355 & 0.148\\
& Sub & 0.289 & 0.435 & \textbf{0.127} & 0.475 & 0.19 & \textbf{0.111} & \textbf{0.433} & 0.361 & \textbf{0.245}\\
& Sub+S & \textbf{0.312} & 0.443 & 0.014 & \textbf{0.498} & \textbf{0.247} & 0.074 & \textbf{0.433} & \textbf{0.368} & \textbf{0.245}\\
\midrule
\multirow{3}{*}{LT} & Acous.  & 0.293 & \textbf{0.304} & 0.005 & 0.356 & 0.291 & 0.0 & 0.359 & \textbf{0.484} & 0.0 \\
& Sub & 0.293 & 0.266 & 0.011 & \textbf{0.393} & 0.278 & 0.0 & \textbf{0.484} & 0.42 & 0.0\\
& Sub+S & \textbf{0.365} & 0.254 & \textbf{0.111} & 0.377 & \textbf{0.305} & 0.0 & 0.459 & 0.382 & 0.0\\
\bottomrule
\end{tabular}
\caption {AQWV scores on ANALYSIS set. +S indicates model with syntactic features.}
\label{aqwvanalysis}
\end{table*}

\section{Experiments and Results}

\paragraph{Pipeline Components} 
All pipeline components were developed by participants in the MATERIAL program \cite{5day}. 
We use the ASR system developed jointly by the University of Cambridge and the University of Edinburgh \cite{ragni18inthewild,carmantini19semisup}.

We evaluate with three different MT systems. We use the neural MT model developed by the University of Edinburgh  (EDI-NMT) \cite{mariannmt} and
the neural and phrase-based statistical MT systems from the University of 
Maryland, UMD-NMT and UMD-SMT respectively \cite{niu-etal-2018-bi}.

For the IR system, we use the bag-of-words query model
implemented in Indri \cite{strohman2005indri}. 

\subsection{Intrinsic Evaluation}
 
We evaluate the models on F-measure of the boundary prediction labels,
as well as WindowDiff \cite{windowdiff}, a 
metric that penalizes difference in the number of boundaries between the reference and predicted segmentation given a fixed window.
We obtain a reference segmentation as described in \autoref{sec:seg-data}.
We indicate our models without and with syntactic features as Sub and Sub+S respectively.
\autoref{intrinsic} shows our results on the ANALYSIS data.
For BG, which is trained on an order of magnitude more data, the model without syntactic information performs slightly better.
Meanwhile, in the lower-data LT setting, 
adding syntactic cues yields a 2.2 point improvement on WindowDiff.

\subsection{Extrinsic Evaluations}

We perform several extrinsic evaluations using 
a pipeline of ASR, ASR segmentation, MT, and information retrieval (IR) components.

\subsubsection{MT Evaluation}
Our first extrinsic evaluation measures the BLEU \cite{papineni2002bleu} score of 
the MT output on the ANALYSIS sets, where we have ground truth 
reference English translations. As our baseline, we compare the same pipeline using the segmentation produced by the acoustic model of the ASR system,
denoted Acous.

Since each segmentation model  produces segments with
different  boundaries, we are unable to use
BLEU directly to compare to the reference sen-
tences.  Thus, we concatenate all segments of a
document and treat them as one segment, which we refer to as ``document-level'' BLEU score.

\autoref{bleu} shows our results.

For BG, both Sub and Sub+S models improve BLEU scores over the baseline
segmentation on the more informal domains (TB, CS). Across all %
MT systems, Sub+S performs best on conversations (CS), 
while Sub performs best on topical monologues
(TB). 

For LT, the segmentation models do not provide any improvement on BLEU scores. However, there is generally an increase in BLEU with the  syntactic features, consistent with the intrinsic results.

\subsubsection{Document-level CLIR Evaluation} Our second extrinsic evaluation is done on the MATERIAL CLIR task.
We are given English queries and asked to retrieve conversations in 
either BG or LT. In our setup, we only search over the English translations
produced by our pipeline. We evaluate the performance of CLIR using 
the Actual Query Weighted Value (AQWV) \cite{aqwv}.

\begin{table*}[ht!]
\centering
\begin{tabular}{ llccc|ccc|ccc }
\toprule
 \multirow{2}{*}{Lang.} & \multirow{2}{*}{Model} & \multicolumn{3}{c}{EDI-NMT} & \multicolumn{3}{c}{UMD-NMT}  & \multicolumn{3}{c}{UMD-SMT} \\
 & & NB & TB & SMT & CS & TB & CS & NB1 & TB & CS\\
\midrule
\multirow{3}{*}{BG} & Acous. & 0.583 & 0.258 & 0.065 & \textbf{0.716} & \textbf{0.305} & \textbf{0.075} & 0.725 & 0.312 & \textbf{0.139}\\
& Sub & \textbf{0.774} & \textbf{0.266} & 0.071 & 0.658 & 0.296 & 0.037 & \textbf{0.675} & 0.383 & 0.076\\
& Sub+S & \textbf{0.774} & 0.186 & \textbf{0.074} & 0.658 & 0.273 & 0.054 & \textbf{0.675} & \textbf{0.407} & 0.105\\
\midrule
\multirow{3}{*}{LT} & Acous.  & 0.161 & 0.195 & 0.262 & \textbf{0.325} & 0.307 & 0.19 & \textbf{0.372} & \textbf{0.404} & 0.262 \\
& Sub & \textbf{0.348} & 0.314 & \textbf{0.333} & 0.271 & 0.385 & \textbf{0.262} & 0.304 & 0.386 & 0.262\\
& Sub+S & 0.269 & \textbf{0.317} & \textbf{0.333} & \textbf{0.300} & \textbf{0.390} & \textbf{0.262} & 0.320 & 0.385 & 0.262 \\
\bottomrule
\end{tabular}
\caption {AQWV scores on DEV set. +S indicates model with syntactic features.}
\label{devclir}
\end{table*}

\begin{table}[ht]
\centering
\begin{tabular}{ c c ccc }
\toprule
\multirow{2}{*}{Lang.} & \multirow{2}{*}{MT} &  \multicolumn{2}{c}{Relevance}  \\
&   & A & M \\
  \midrule
  \multirow{3}{*}{BG} 
 & EDI-NMT &   0.564 & \textbf{0.566}\\
 & UMD-NMT &  0.572 &\textbf{0.615} \\
   & UMD-SMT & 0.593 &  \textbf{0.658}\\
  & Reference & \multicolumn{2}{c}{0.862} \\
   \midrule
\multirow{3}{*}{LT} & EDI-NMT  & \textbf{0.576} & 0.554  \\
 & UMD-NMT  & \textbf{0.663} & 0.593  \\
  & UMD-SMT  & \textbf{0.681} & 0.614 \\
  & Reference & \multicolumn{2}{c}{0.9}  \\
\bottomrule
\end{tabular}
\caption {Passage-level evaluation comparing relevance using the Sub+S model (M), the acoustic baseline (A).Evaluation of reference translation is also provided for each language.}
\label{human}

\end{table}

\autoref{aqwvanalysis} shows the results of the CLIR ANALYSIS evaluation. 

Similar trends are found on the DEV set.
On BG, our models yield large increases in AQWV for both UMD MT models,
especially on CS, where the gains are on the order of 0.1 absolute points.
Syntactic information also proves useful, as Sub+S performs best 
in six of nine settings.
Despite the lack of increase in BLEU for LT, the segmentation models show large increases in AQWV over the baseline, especially on UMD-SMT/NB where the Sub model improves AQWV by 0.125 points absolutely.
Only EDI-NMT was able to yield nonzero retrieval scores for the CS 
domain, with Sub+S improving by 0.106 points.

\subsubsection{Passage-level CLIR Evaluation}
We also conduct a human evaluation to compare our segmentation model with 
acoustically-based segmentation and investigate 
which makes it easier for %
annotators to determine MT quality and query relevance. To this end, we collect 
relevant query/passage pairs and ask Amazon Mechanical Turk Workers\footnote{\url{https://www.mturk.com/}}
to judge quality and relevance. 

The MT quality judgments were done on a 7-point scale (integer scores from -3 to 3 inclusive), and the query relevance judgments on a 3-point scale (0, 0.5, and 1). A perfect pipeline should achieve 3 in MT quality and 1 in query relevance. We give each HIT (each containing five passages) to three distinct Workers. \autoref{amtinstruction} shows the detailed instruction we have for the HIT. Also see \autoref{amthit} for an example passage as displayed in a HIT.

We require Workers to have a minimum lifetime approval rate of 98\% and number of HITs approved greater than 5000. Workers that provide the same quality score for all snippets in a HIT are manual checked by the author.

To generate our evaluation data, 
we use YAKE! \cite{campos2018text} to extract keywords from documents in the ANALYSIS dataset. 
We then collect 3-segment passages of each document and pair them with a keyword that appears in the middle utterance in the ground truth transcription; these will become the passages and queries we give to Workers. 
We match the timestamps of these passages in the ground truth transcription with the output of the Sub+S model and the Acous. model,
and feed those segments through MT. We randomly sample 200 passages each from BG and LT
and present them in three conditions (ground truth or pipeline with either our segmentation, or acoustic segmentation).
Please refer to \autoref{amtinstruction} and \autoref{amthit} in the appendix for the instruction and example provided for the Mechanical Turk task.

\autoref{human} shows the results. %
We omit the differences in quality because they were not significant. 
The human reference
transcriptions received 0.917 (BG) and  1.153 (LT) out of a maximum
of 3.0, suggesting that speech excerpts, even when well translated, are hard to understand out of context. 
On the relevance assessment, we see consistent improvements in BG using the Sub+S model, regardless of the MT system, although only UMD-SMT is statistically significant.\footnote{Using the approximate randomization test at the $\alpha=.05$ level \cite{riezler-maxwell-2005-pitfalls}.}

We do not see improvements on relevance on LT, although no differences
are significant. While this might seem counter-intuitive, given 
that the Sub+S model leads to consistent improvement in LT CLIR, it is corroborated by the lower BLEU scores on LT, suggesting the CLIR pipeline is less affected by poor fluency than are actual human users who need to read the output to determine relevance.

\subsection{Discussion}
Overall, when subtitle data is plentiful, as is the case with BG, we 
see consistent improvements on downstream MT and CLIR tasks. Moreover, 
we find consistent improvements in the CS domain where acoustic segmentation is likely to produce choppy, non-sentence-like segments.
Even on LT, where there is not enough data to realize gains in translation,
it still has positive effects on the document-level CLIR task.

\section{Conclusion}
We present an approach for ASR segmentation for low-resource languages for the task of STTT. On extrinsic evaluations of MT, IR, and human evaluations, we are able to show improvements in the downstream MT and CLIR. In future work, we hope to explore methods to make the 
tagger model more robust to noise, since word-error rates of ASR
in the low-resource condition tend to be high

\section{Acknowledgements}
This research is based upon work supported inpart by the Office of the Director of National Intelligence (ODNI), Intelligence Advanced Research Projects Activity (IARPA), via contract \#FA8650-17-C-9117. The views and conclusions contained herein are those of the authors and should not be interpreted as necessarily representing the official policies, either expressed or implied, of ODNI, IARPA, or the U.S. Government. The U.S. Government is authorized to reproduce and distribute reprints for governmental purposes not withstanding any copyright annotation therein.

\section{References}\label{reference}

\bibliographystyle{lrec}
\bibliography{segmentation}

\appendix

\section{Additional Document-level CLIR Evaluation}
\label{devaqwv}

\begin{figure}[h]
    \centering
    \includegraphics[width=.5 \textwidth]{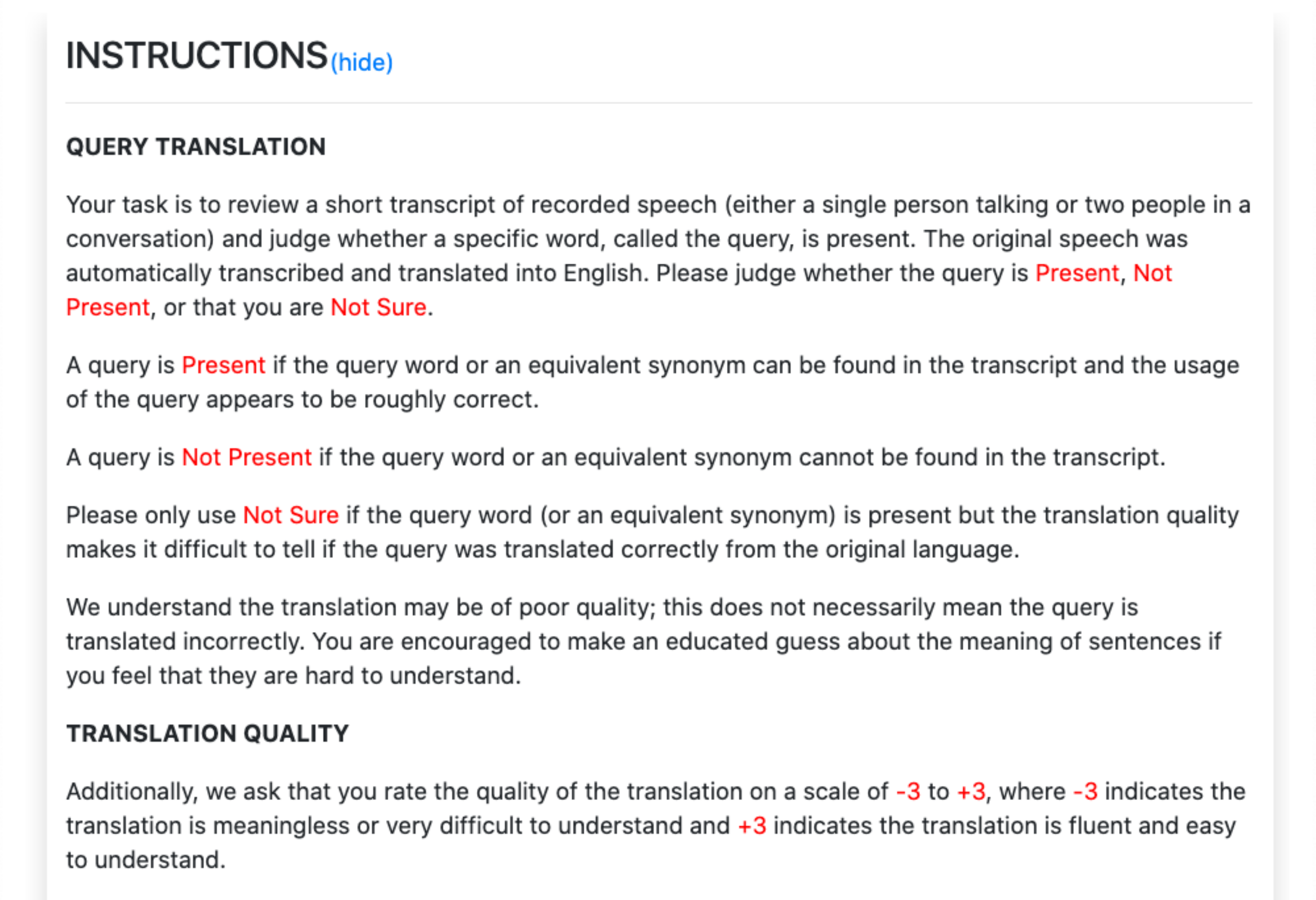}
    \caption{The instructions we provided for the Mechanical Turk task.}
    \label{amtinstruction}
\end{figure}

\begin{figure}[h]
\centering
\includegraphics[width=.5 \textwidth]{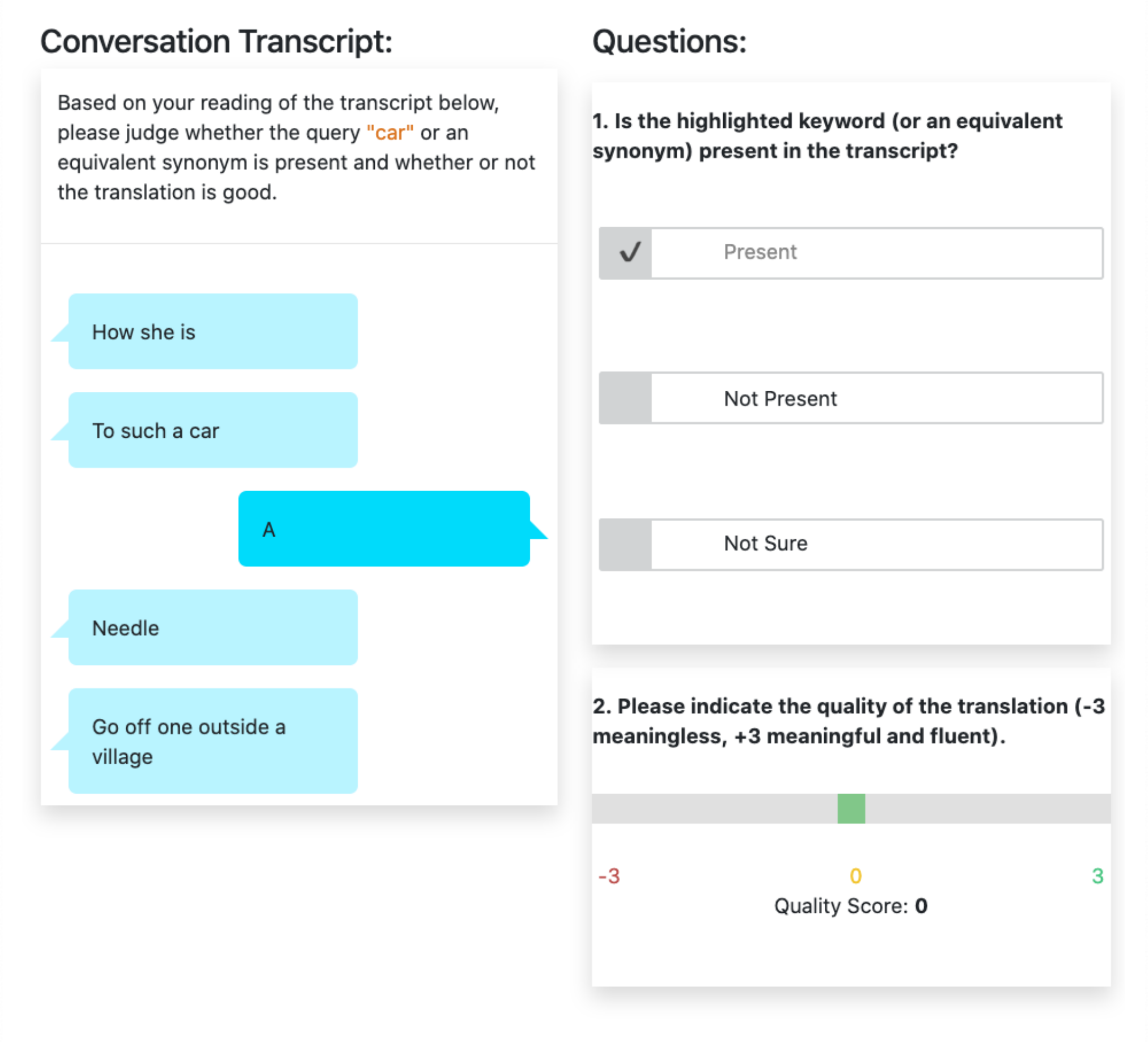}
\caption{An example of our Amazon Mechanical Turk relevance and quality judgment task.}
\label{amthit}
\end{figure}
\clearpage

\end{document}